\documentclass[journal]{IEEEtran}
\makeatletter\if@twocolumn\PassOptionsToPackage{switch}{lineno}\else\fi\makeatother

      \makeatletter
\usepackage{wrapfig}
\newcounter{aubio}

\usepackage{algorithm}
\usepackage{algpseudocode}
\usepackage{tabularx}
\usepackage{amsmath,amsfonts,amsthm,amssymb,amsopn,bm}

\long\def\bioItem{%
\@ifnextchar[{\@bioItem}{\@@bioItem}}

\long\def\@bioItem[#1]#2#3{
 \stepcounter{aubio}
 \expandafter\gdef\csname authorImage\theaubio\endcsname{#1}
 \expandafter\gdef\csname authorName\theaubio\endcsname{#2}
 \expandafter\gdef\csname authorDetails\theaubio\endcsname{#3}
}

\long\def\@@bioItem#1#2{
 \stepcounter{aubio}
 \expandafter\gdef\csname authorName\theaubio\endcsname{#1}
 \expandafter\gdef\csname authorDetails\theaubio\endcsname{#2}
}

\newcommand{\checkheight}[1]{%
  \par \penalty-100\begingroup%
  \setbox8=\hbox{#1}%
  \setlength{\dimen@}{\ht8}%
  \dimen@ii\pagegoal \advance\dimen@ii-\pagetotal
  \ifdim \dimen@>\dimen@ii
    \break
  \fi\endgroup}

\def\printBio{%
  \@tempcnta=0
   \loop
     \advance \@tempcnta by 1
     \def\aubioCnt{\the\@tempcnta}
     \setlength{\intextsep}{0pt}%
     \setlength{\columnsep}{10pt}%
     \expandafter\ifx\csname authorImage\aubioCnt\endcsname\relax%
      \else%
       \checkheight{\includegraphics[height=1.25in,width=1in,keepaspectratio]{\csname authorImage\aubioCnt\endcsname}}
        \begin{wrapfigure}{l}{25mm}
         \includegraphics[height=1.25in,width=1in,keepaspectratio]{\csname authorImage\aubioCnt\endcsname}
        \end{wrapfigure}\par
      \fi
     \noindent\textbf{\csname authorName\aubioCnt\endcsname}\csname authorDetails\aubioCnt\endcsname \par\bigskip
      \ifnum\@tempcnta < \theaubio
   \repeat
   }
\makeatother

\usepackage{graphicx}
\usepackage[T1]{fontenc}
\ifCLASSINFOpdf
\else
\fi
\usepackage{amsmath}
\usepackage[cmintegrals]{newtxmath}
\usepackage{url,multirow,morefloats,floatflt,cancel,tfrupee}
\makeatletter

\AtBeginDocument{\@ifpackageloaded{textcomp}{}{\usepackage{textcomp}}}
\makeatother
\usepackage{colortbl}
\usepackage{xcolor}
\usepackage{pifont}
\usepackage[nointegrals]{wasysym}
\makeatletter

\def\mcWidth#1{\csname TY@F#1\endcsname+\tabcolsep}

\def\cAlignHack{\rightskip\@flushglue\leftskip\@flushglue\parindent\z@\parfillskip\z@skip}
\def\rAlignHack{\rightskip\z@skip\leftskip\@flushglue \parindent\z@\parfillskip\z@skip}

\usepackage{ifxetex}
\ifxetex\else\if@twocolumn\usepackage{dblfloatfix}\fi\fi

\AtBeginDocument{
\expandafter\ifx\csname eqalign\endcsname\relax
\def\eqalign#1{\null\vcenter{\def\\{\cr}\openup\jot\m@th
  \ialign{\strut$\displaystyle{##}$\hfil&$\displaystyle{{}##}$\hfil
      \crcr#1\crcr}}\,}
\fi
}

\AtBeginDocument{%
  \@ifpackageloaded{endfloat}%
   {\renewcommand\efloat@iwrite[1]{\immediate\expandafter\protected@write\csname efloat@post#1\endcsname{}}}{\newif\ifefloat@tables}%
}%

\def\BreakURLText#1{\@tfor\brk@tempa:=#1\do{\brk@tempa\hskip0pt}}
\let\lt=<
\let\gt=>
\def\processVert{\ifmmode|\else\textbar\fi}

\@ifundefined{subparagraph}{
\def\subparagraph{\@startsection{paragraph}{5}{2\parindent}{0ex plus 0.1ex minus 0.1ex}%
{0ex}{\normalfont\small\itshape}}%
}{}

\newcommand\role[1]{\unskip}
\newcommand\aucollab[1]{\unskip}
  
\@ifundefined{tsGraphicsScaleX}{\gdef\tsGraphicsScaleX{1}}{}
\@ifundefined{tsGraphicsScaleY}{\gdef\tsGraphicsScaleY{.9}}{}
\def\checkGraphicsWidth{\ifdim\Gin@nat@width>\linewidth
	\tsGraphicsScaleX\linewidth\else\Gin@nat@width\fi}

\def\checkGraphicsHeight{\ifdim\Gin@nat@height>.9\textheight
	\tsGraphicsScaleY\textheight\else\Gin@nat@height\fi}

\def\fixFloatSize#1{}
\let\ts@includegraphics\includegraphics

\def\inlinegraphic[#1]#2{{\edef\@tempa{#1}\edef\baseline@shift{\ifx\@tempa\@empty0\else#1\fi}\edef\tempZ{\the\numexpr(\numexpr(\baseline@shift*\f@size/100))}\protect\raisebox{\tempZ pt}{\ts@includegraphics{#2}}}}

\AtBeginDocument{\def\includegraphics{\@ifnextchar[{\ts@includegraphics}{\ts@includegraphics[width=\checkGraphicsWidth,height=\checkGraphicsHeight,keepaspectratio]}}}

\DeclareMathAlphabet{\mathpzc}{OT1}{pzc}{m}{it}

\def\URL#1#2{\@ifundefined{href}{#2}{\href{#1}{#2}}}

\def\UrlOrds{\do\*\do\-\do\~\do\'\do\"\do\-}%
\g@addto@macro{\UrlBreaks}{\UrlOrds}

\@ifundefined{quoteAttrib}
	{}
	{}

\@ifundefined{titlequoteAttrib}
	{}{}

\newenvironment{title-quote}
	{\list{}{\fontsize{10pt}{12pt}\selectfont\leftmargin.5in\itshape\rightmargin\leftmargin}%
  \item\relax}
  {\endlist}

\makeatother

\usepackage{tabulary}
\makeatletter
\AtBeginDocument{\@ifpackageloaded{longtable}{%
\def\LT@makecaption#1#2#3{%
  \LT@mcol\LT@cols c{\hbox to\z@{\hss\parbox[t]\LTcapwidth{%
    \sbox\@tempboxa{#1{#2: } #3}%
    \ifdim\wd\@tempboxa>\hsize
      #1{#2: }\textsc{#3}%
    \else
      \hbox to\hsize{\hfil\box\@tempboxa\hfil}%
    \fi
    \endgraf\vskip\baselineskip}%
  \hss}}}
}{}}
\makeatother

   \makeatletter
  \def\fig@textbf{\textbf}
   \AtBeginDocument{\renewcommand\floatc@plain[2]{\setbox\@tempboxa\hbox{{\footnotesize#1.}\footnotesize\hskip.5em#2}%
    \ifdim\wd\@tempboxa>\hsize {\fig@textbf{\footnotesize#1.}}\footnotesize\hskip.5em#2\par
        \else\hbox to\hsize{\hfil\box\@tempboxa\hfil}\fi}}
    \makeatother
  
\usepackage{float}

\begin{document}

%

        \title{\textbf{Safe, Efficient, and Comfortable Velocity Control based on Reinforcement Learning for Autonomous Driving}}
      
\author{Meixin~Zhu,
        Yinhai~Wang,
        Ziyuan Pu,
        Jingyun~Hu,
        Xuesong~Wang, and 
        Ruimin~Ke \thanks{Meixin~Zhu, Yinhai~Wang, Xuesong~Wang are with School of Transportation Engineering, Tongji University, Shanghai, 201804, China}\thanks{Meixin~Zhu, Yinhai~Wang, Ziyuan Pu, Jingyun~Hu, Ruimin~Ke are with Department of Civil and Environmental Engineering, University of Washington, Seattle, 98195, Washington, United States, e-mail: meixin92@uw.edu, yinhai@uw.edu, ziyuanpu@uw.edu, jingyun@uw.edu, ker27@uw.edu}\thanks{Xuesong~Wang is with Key Laboratory of Road and Traffic Engineering, Ministry of Education, Tongji University, Shanghai, China, e-mail: wangxs@tongji.edu.cn}\thanks{Corresponding author: 
        Yinhai~Wang and Xuesong Wang}}

\maketitle 


\IEEEpeerreviewmaketitle

\textbf{ABSTRACT}

A model used for velocity control during car following was proposed based on deep reinforcement learning (RL). To fulfill the dual objectives of imitating human drivers and optimizing driving performance, a reward function was developed by referencing human driving data and combining driving features related to safety, efficiency, and comfort. With the reward function, the RL agent learns to control vehicle speed in a fashion that maximizes cumulative rewards, through trials and errors in the simulation environment. A total of 1,341 car-following events extracted from the Next Generation Simulation (NGSIM) dataset were used to train the model. Car-following behavior produced by the model were compared with that observed in the empirical NGSIM data, to demonstrate the model's ability to follow a lead vehicle safely, efficiently, and comfortably. Results show that the model demonstrates the capability of safe, efficient, and comfortable velocity control in that it 1) has small percentages (8\%) of dangerous minimum time to collision values (\textless\ 5s) than human drivers in the NGSIM data (35\%); 2) can maintain efficient and safe headways in the range of 1s to 2s; and 3) can follow the lead vehicle comfortably with smooth acceleration. The results indicate that proposed approach could contribute to the development of better autonomous driving systems.\\

\textit{Keywords:} Car Following, Autonomous Driving, Velocity Control, Reinforcement Learning, NGSIM, Deep Deterministic Policy Gradient (DDPG)
    
\section{INTRODUCTION}
Car following is the most frequent driving scenario. The main task of car following is controlling vehicle velocity to keep safe and comfortable following gaps. Autonomous car-following velocity control has the promise to mitigate drivers' workload, to improve traffic safety, and to increase road capacity\unskip~\cite{369980:8164317,zhu2018modeling}. 

Driver models are critical elements of velocity control systems \cite{wang2016drivers,wang2016development}. In general, driver models related to car following have been established with two approaches: rule-based and supervised learning\unskip~\cite{369980:8164318, feiyue}. Rule-based approach mainly refers to traditional car-following models, such as the Gaxis-Herman-Rothery model\unskip~\cite{369980:8164259} and the intelligent driver model\unskip~\cite{369980:8164320}. Supervised learning approach relies on data typically provided through human demonstration in order to approximate the relationship between car-following state and acceleration.

These two approaches all intend to emulate human drivers' car-following behavior. However, solely imitating human driving behaviors may not be the best solution in autonomous driving. Firstly, human drivers may not drive in an optimal way \cite{chai2015fuzzy}. Secondly, users may not want their autonomous vehicles driving in a way like them\unskip~\cite{369980:8164261}. Thirdly, driving should be optimized with respect to safety, efficiency, comfort, besides imitating human drivers.

To resolve the problem, we propose a car-following model for autonomous velocity control based on deep reinforcement learning (RL). This model not only tries to emulate human drivers but also directly optimizes driving safety, efficiency, and comfort, by learning from trial and interaction with a simulation environment. 

Specifically, the deep deterministic policy gradient (DDPG) algorithm\unskip~\cite{369980:8164262} that performs well in continuous control field was utilized to learn an actor network together with a critic network.  The actor is responsible for policy generation: outputting following vehicle accelerations based on speed, relative speed, and spacing. The critic is responsible for policy improvement: update the actor's policy parameters in the direction of performance improvement. 

To evaluate the proposed model, real-world driving data collected in the Next Generation Simulation (NGSIM) project\unskip~\cite{369980:8164263} were used to train the model. And car-following behavior simulated by the DDPG model was compared with that observed in the empirical NGSIM data, to demonstrate the model's ability to follow a leading vehicle safely, efficiently, and comfortably.
    
\section{BACKGROUND}

\subsection{Car Following}
Car-following models describe the movements of a following vehicle (FV) in response to the actions of the lead vehicle (LV) \cite{zhu2018modeling}. They are essential components of microscopic traffic simulation \cite{brackstone1999car}, and serve as theoretical references for autonomous car-following systems \cite{Wei}. Since the early investigation of car-following dynamics in 1953 \cite{pipes1953operational}, numerous car-following models have been built.

The first car-following model \cite{pipes1953operational} was proposed in the middle 1950s, and a number of models have been developed since then, for example, the Gaxis-Herman-Rothery (GHR) model \cite{gazis1961nonlinear}, the intelligent driver model (IDM) \cite{treiber2000congested}, the optimal velocity model \cite{bando1995dynamical}, and the models proposed by Helly \cite{helly1959simulation}, Gipps \cite{gipps1981behavioural}, and Wiedemann \cite{wiedemann1974simulation}. For detailed review and historical development of the subject, consult Brackstone and McDonald \cite{brackstone1999car} and Saifuzzaman and Zheng \cite{saifuzzaman2014incorporating}.

\subsection{Reinforcement Learning}Reinforcement learning optimizes sequential decision-making problems by letting an RL agent interact with an environment. At time step \textit{t}, the agent observes a state \textit{s\ensuremath{_{t}}} and chooses an action \textit{a\ensuremath{_{t}}} from some action space \textit{A} based on a policy \textit{\ensuremath{\pi }}(\textit{a\ensuremath{_{t}}}|\textit{s\ensuremath{_{t}}}) that maps from state \textit{s\ensuremath{_{t}}} to actions \textit{a\ensuremath{_{t}}}. Meanwhile, the system gives a reward \textit{r\ensuremath{_{t}}} to the agent, and transits to the next state \textit{s\ensuremath{_{t+}}}\ensuremath{_{1}}. This process continues until a terminal state is reached, then the agent restarts. The agent intends to get a maximum discounted, accumulated reward $R _ { t } = \sum _ { k = 0 } ^ { \infty } \gamma ^ { k } r _ { t + k }$, with the discount factor $\gamma \in ( 0,1 ]$ \cite{369980:8164264}. In general, there are two types of RL methods: value-based and policy-based\unskip~\cite{369980:8164265}.

\subsubsection{Value-Based Reinforcement Learning}A value function measures the quality of a state or state state-action pair. The action value $Q ^ { \pi } ( s , a ) = E \left[ R _ { t } | s _ { t } = s , a _ { t } = a \right]$ is the expected return for selecting action \textit{a} in state \textit{s} and then following policy \textit{\ensuremath{\pi }}. It represents the goodness of taking action \textit{a} in a state \textit{s}. Value-based RL methods intend to infer the action value function from historical experience. $Q$-learning is a typical value-based RL method. Beginning with a random $Q$-function, the agent keeps updating its $Q$-values based on the Bellman equation\unskip~\cite{369980:8164265}.
\begin{equation} \label{eq:bellman}
Q ( s , a ) = E \left[ r + \gamma \max _ { a ^ { \prime } } Q \left( s ^ { \prime } , a ^ { \prime } \right) \right]
\end{equation}

The intuition is that: maximum future reward for this state \textit{s} and action \textit{a} is the immediate reward \textit{r} plus maximum future reward for the next state $s ^ { \prime }$. Based on the estimated \textit{Q}-values, the optimal policy is to take the action with the highest $Q ( s , a )$ to get maximum expected future rewards.

\subsubsection{Policy-Based Reinforcement Learning}Different with value-based methods, policy-based methods try to improve the policy $\pi ( a | s ; \theta )$ directly, by updating its parameters \textit{\ensuremath{\theta }} with gradient ascent on $E \left[ R _ { t } \right]$. A typical policy-based method is REINFORCE, which updates the policy parameters \textit{\ensuremath{\theta }} with $\nabla _ { \theta } \log \pi \left( a _ { t } | s _ { t } ; \theta \right) R _ { t }$ \unskip~\cite{369980:8164264} . 

To reduce the variance of policy gradients and increase learning speed, an actor-critic method is usually adopted. Two learning agents are used in an actor-critic algorithm: the actor (policy) and the critic (value function). The actor determines which action to take, and the critic tells the actor the quality of the action and how it should adjust the policy\cite{369980:8164266}.

\subsection{Deep Reinforcement Learning}Deep reinforcement learning refers to reinforcement learning algorithms that use neural networks to approximate value function $V ( s ; \theta ) \text {, policy } \pi ( a | s ; \theta )$, or system model.

\subsubsection{Deep Q-Network}Instead of computing $Q(s,a)$ for each state-action pair, deep $Q$-learning uses neural network as function approximator to estimate the action-value function \cite{369980:8164267}. The action is selected with a maximum $Q(s,a)$ value. Deep \textit{Q} networks (\textit{DQN}) work well with discrete action spaces but fail in continuous action spaces, like in our case. To address this, Lillicrap et al.\unskip~\cite{369980:8164262} developed an algorithm called deep deterministic policy gradient (DDPG). DDPG introduced an actor-critic mechanism to \textit{DQN } and can be used for continuous control problems.

\subsubsection{Deep Deterministic Policy Gradient}DDPG uses two separate networks to approximate the actor and critic respectively \cite{369980:8164262}. The critic network with weights \textit{\ensuremath{\theta }\ensuremath{^{Q }}} is responsible for estimating the action-value function \textit{Q}(\textit{s,a|\ensuremath{\theta }\ensuremath{^{Q}}}). The actor network with weights \textit{\ensuremath{\theta }\ensuremath{^{\mu }}} is responsible for explicitly representing the agent's policy \textit{\textmu }(\textit{s|\ensuremath{\theta }\ensuremath{^{\mu }}}). As proposed in $DQN$, experience replay and target network are adopted in DDPG to facilitate stable and robust learning.

  \begin{itemize}
  \item \relax Experience replay
  
  A replay buffer was applied to avoid learning from sequentially generated, correlated experience samples. The replay buffer is a finite sized cache \textit{D} that stores transitions (\textit{s\ensuremath{_{t}}}, \textit{a\ensuremath{_{t}}, r\ensuremath{_{t}}, s\ensuremath{_{t+}}}\ensuremath{_{1}}) sampled from the environment. The replay buffer is continually updated by replacing old samples with new ones. At each time step, the actor and critic networks are trained on random mini-batches of transitions from the replay buffer.

  \item \relax Target network
  
  Target networks are used to represent target values of the main networks, to avoid divergence of the algorithm \cite{369980:8164267}.
  Two target networks, $Q ^ { \prime } ( s , a | \theta ^ { Q^\prime } ) \text { and } \mu ^ { \prime } ( s | \theta ^ { \mu ^ { \prime } } )$, were created for the main critic and actor networks respectively. They have the same architecture with the main networks but with different network parameters $\theta^\prime$. The parameters of target networks are updated by letting them slowly track the main networks: $\theta ^ { \prime } = \tau \theta + ( 1 - \tau ) \theta ^ { \prime } \text { with } \tau \ll 1$. In this way, the target values are constrained to update slowly, greatly enhancing the stability of learning.
    \end{itemize}

The full DDPG algorithm is listed in Algorithm I. It begins with initializing the replay buffer and the actor, critic and corresponding target networks. At each time step, an action $a$ is taken according to the exploratory policy. Then, the reward $r_t$ and new state $s_{t+1}$ are observed and stored in the replay memory \textit{D}. The critic is trained with mini-batches sampled from the replay memory. Afterward, the actor is updated by performing a gradient ascent step on the sampled policy gradient. Finally, the target networks with weights $\theta^{Q^\prime}$  and $\theta^{\mu^\prime}$ are updated to slowly track the actor and critic networks. 

\begin{algorithm*}[!htbp]\label{alg1}
\ignorespaces 
\caption{DDPG: Deep deterministic policy gradient for car-following velocity control}\label{euclid}
\begin{algorithmic}[1]
\State Randomly initialize critic $Q(s,a|{\theta ^Q})$ and actor $\mu (s|{\theta ^\mu })$ networks with weights ${\theta ^Q}$ and ${\theta ^\mu }$.
\State Initialize target network $Q'(s,a|{\theta ^{Q'}})$ and $\mu '(s|{\theta ^{\mu '}})$ with weights ${\theta ^{Q'}} \leftarrow {\theta ^Q}$ and ${\theta ^{\mu '}} \leftarrow {\theta ^\mu }$
\State Set up empty replay buffer D
\For  {episode = 1 to M}
\State Begin with a random process N for action exploration
\State Observe initial car-following state: initial gap, follower speed, and relative speed
\For  {episode = 1 to T}
\State Calculate reward $r_{t}$
\State Choose follower acceleration $a_t=\mu(s_t│\theta^{\mu} )+N_t$ based on current actor network and exploration noise $N_t$
\State Implement acceleration $a_{t}$ and transfer to new state $s_{t+1}$ based on kinematic point-mass model
\State Save transition $(s_{t}, a_{t}, r_{t}, s_{t+1})$ into replay buffer $D$
\State Sample random minibatch of $N$ transitions $(s_{i}, a_{i}, r_{i}, s_{i+1})$ from $D$
\State Set ${y_i} = {r_i} + \gamma Q'({s_{i + 1}},\mu '({s_{i + 1}}|{\theta ^{\mu '}})|{\theta ^{Q'}})$
\State Update critic through minimizing loss: 
$L = \frac{1}{N}\sum\nolimits_i {{{({y_i} - Q({s_i},{a_i}|{\theta ^Q}))}^2}} $
\State 
Update actor policy using sampled policy gradient: 
${\nabla _{{\theta ^\mu }}}J \approx \frac{1}{N}\sum\limits_i {{\nabla _a}Q(s,a|{\theta ^Q}){|_{s = {s_i},a = \mu ({s_i})}}{\nabla _{{\theta ^\mu }}}\mu (s|{\theta ^\mu })} {|_{{s_i}}}$
\State Update target networks:$\begin{array}{l} {\theta ^{Q'}} = \tau {\theta ^Q} + (1 - \tau ){\theta ^{Q'}}\\
{\theta ^{\mu '}} = \tau {\theta ^\mu } + (1 - \tau ){\theta ^{\mu '}}
\end{array}$ 
\EndFor
\EndFor
\end{algorithmic}
\end{algorithm*}


\section{DATA PREPARATION}
\begin{figure}[h]
    \centering
    \includegraphics[width=0.45\textwidth]{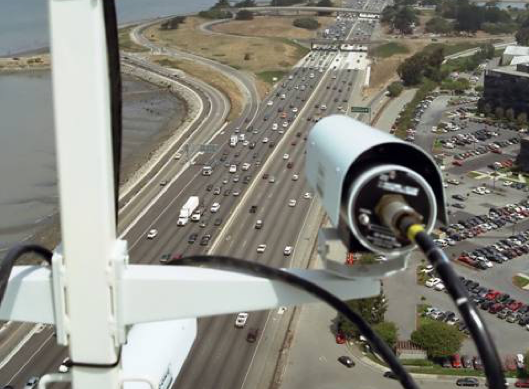}
    \caption{The layout of the road segment studied in dataset.}
    \label{fig:ngsim}
\end{figure}

Vehicle trajectory data in the Next Generation Simulation (NGSIM) project\unskip~\cite{369980:8164263}  were used. Specifically, this study used for data retrieved from eastbound I-80 in the San Francisco Bay area in Emeryville, CA, on April 13, 2005, as shown in Fig. \ref{fig:ngsim}. The investigation region was around 500 meters (1,640 feet) long and comprised of six freeway lanes, including a high-occupancy vehicle (HOV) lane. An aggregate of 45 minutes of data are accessible in the full dataset, divided into three 15-minute time spans: 4:00 p.m. to 4:15 p.m.; 5:00 p.m. to 5:15 p.m.; and 5:15 p.m. to 5:30 p.m. These periods contain the congestion buildup, or the inter-state between uncongested and congested traffic states, and full congestion during a peak period. The data provide the precise location information for each vehicle, with the sampling rate being 10 Hz. To enhance data quality, the reconstructed NGSIM I-80 data\unskip~\cite{369980:8164268} were utilized.

Car-following events were extracted by applying a car-following filter as described in Wang et al. \cite{wang2018capturing}. A car-following event was defined as:

\begin{itemize}
  \item \relax The leading and following vehicle pairs stay in the same lane;
  
  \item \relax Duration of the event {\textgreater} 15s: ensuring that the car-following persisted long enough to be analyzed.
  \end{itemize}
  A total of 1,341 car-following events were extracted and utilized in this study.

\section{FEATURES FOR REWARD FUNCTION}\label{feature}
In this section, features that capture relevant objectives of the car- following velocity control were proposed, with a final aim to construct a proper reward function.

\subsection{Safety}Safety should be the most important element of autonomous car following. Time to collision (TTC) was used to represent safety. As a widely used safety indicator, TTC represents the time left before two vehicles collide. It is computed as:
\begin{equation} \label{eq:ttc}
T T C ( t ) = \frac { S _ { n - 1 , n } ( t ) } { \Delta V _ { n - 1 , n } ( t ) }
\end{equation}
where $S_{ n - 1 , n}$ is the following gap, $\Delta V _ { n - 1 , n }$ is the relative speed.

\begin{figure}[h]
    \centering
    \includegraphics[width=0.45\textwidth]{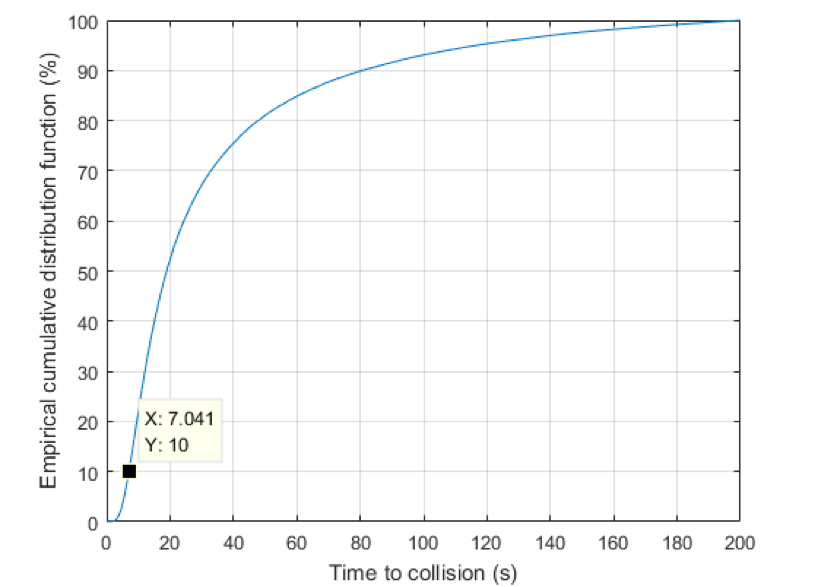}
    \caption{Cumulative distribution of TTC in car-following events extracted from the NGSIM.}
    \label{fig:ttc}
\end{figure}

TTC is inversely related to crash risk (smaller TTC values correspond to higher crash risks and vice versa)\unskip~\cite{369980:8164269}. To apply TTC as a feature reflecting safety, a safety limit (a lower bound of TTC) should be determined. However, different thresholds (from 1.5s to 5s) are reported in the literature \cite{369980:8164269}. To address this problem, we determine the safety limit based on the NGSIM empirical data. Fig. \ref{fig:ttc} shows the cumulative distribution of TTC values in car-following events extracted from the NGSIM data. A 7-second safety limit corresponding to the 10 percentiles of TTC distribution was chosen. Then the TTC feature was constructed as:
\begin{equation}\label{eq:ttcfeature}
    F_{TTC} = \begin{cases} 
      \log(TTC/7) & 0 \leq TTC \leq 7 \\
      0 & \text{otherwise}
   \end{cases}
\end{equation}
  
In this way, if TTC is less than 7s, the TTC feature will be negative. And as TTC approaches zeros, the TTC feature will be close to negative infinity, which represents a severe punishment to near-crash situations.

\subsection{Efficiency}Time headway was used to measure driving efficiency, and it is defined as the passed time between the arrival of the lead vehicle (LV) and the following vehicle (FV) at a designated point. Keeping a short headway within the safety bounds can improve traffic flow efficiency because short headways correspond to large roadway capacities\unskip~\cite{369980:8164270}. 

The rules of different countries are not quite the same, in regard to the legal or recommended time headway. In the U.S., several driver training programs state that it is difficult to follow a vehicle safely with headway being less than 2 s. In Germany, the recommended time headway is 1.8 s, and fines are imposed when the time headway is less than 0.9 s. In Sweden, the police use a time headway of 1 s as a threshold for imposing fines\unskip~\cite{369980:8164269}.

\begin{figure}[h]
    \centering
    \includegraphics[width=0.45\textwidth]{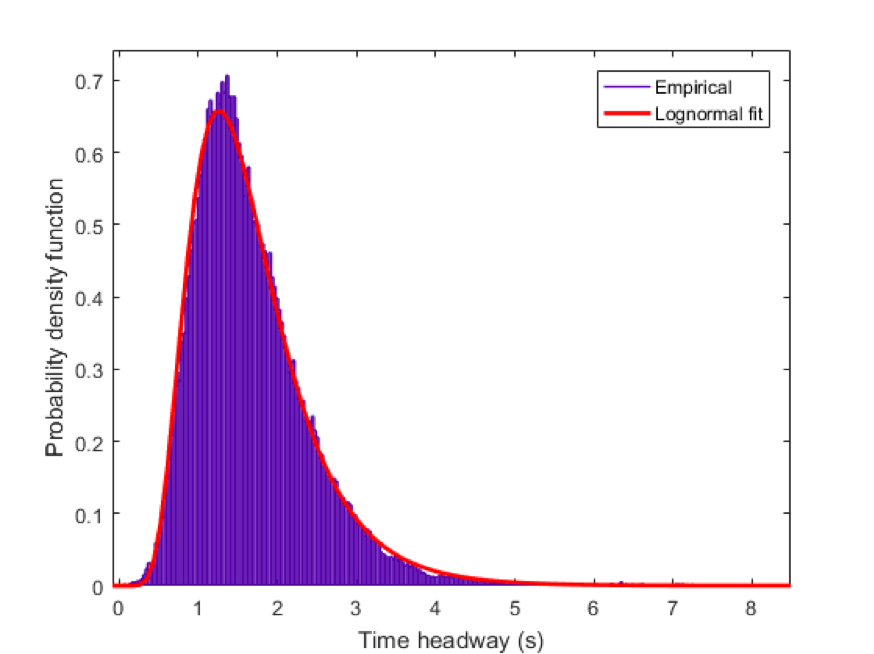}
    \caption{Distribution of time headway in car-following events extracted from the NGSIM data.}
    \label{fig:hdw}
\end{figure}

This study determined the appropriate time headway based on the empirical NGSIM data. Fig. \ref{fig:hdw} presents the distribution of time headway in all of the extracted 1,341 car-following events. A lognormal distribution was fit on the data. The lognormal distribution is a probability distribution whose logarithm has a normal distribution. The probability density function of the lognormal distribution is:
\begin{equation}\label{eq:lognormal}
    f _ { \text { lognorm } } ( x | \mu , \sigma ) = \frac { 1 } { x \sigma \sqrt { 2 \pi } } e ^ { \frac { - ( \ln x - \mu ) ^ { 2 } } { 2 \sigma ^ { 2 } } } ; x > 0
\end{equation}
where $x$ is the distribution variable, time headway in this study, and  $\mu, \sigma$ are the mean and log standard deviation of the variable \textit{x}, respectively. Based on the empirical data, the estimated $\mu$ and $\sigma$ were 0.4226 and 0.4365 respectively.

A headway feature was constructed as the probability density value of the estimated headway lognormal distribution: 
\begin{equation}\label{eq:hdwfeature}
    F _ { h e a d w a y } = f _ { \text { lognorm } } ( h e a d w a y | \mu = 0.4226 , \sigma = 0.4365 )
\end{equation}

According to this headway feature, headways around 1.3 seconds correspond to large headway feature values (about 0.65); while headways being too long or too short correspond to low feature values. In this way, efficient headways are encouraged while unsafe or too long headways are discouraged.

\subsection{Comfort}Jerk, defined as the change rate of acceleration, was used to measure driving comfort because it has a strong influence on the comfort of the passengers\unskip~\cite{369980:8164271}. A jerk feature was constructed as:
\begin{equation}\label{eq:jerkfeature}
    F _ { j e t t } = \frac { {jerk } ^ { 2 } } { 3600 }
\end{equation}

The squared jerk was divided by a base value (3600) to scale the feature into the range of [0 1]. The base value was determined by the following intuition:
  
  \begin{enumerate}
  \item \relax The sample interval of the data is 0.1s;
  \item \relax The acceleration is bounded between -3 to 3 m/s\ensuremath{^{2}} based on the observed FV acceleration of all the car-following events;
  \item \relax Therefore the largest jerk value is $\frac { 3 - ( - 3 ) } { 0.1 } = 60 m / s ^ { 3 }$, if squared, we get 3600.
  \end{enumerate}

\section{PROPOSED APPROACH}
Since vehicle acceleration is a continuous variable, deep deterministic policy gradient (DDPG)\unskip~\cite{369980:8164262} algorithm was used. In this section, the approach proposed to learn velocity control strategy using DDPG is explained.

\subsection{State and Action}At a certain time step \textit{t}, the state of a car-following process is described by the FV speed  $V_{n}(t)$, spacing $S_{n-1,n}(t)$, and relative speed $\Delta V_{n-1,n}(t)$. The action is the longitudinal acceleration of the FV $a_{n}(t)$. Given state and action at time step \textit{t}, the next-step state is updated by a kinematic point-mass model: 
\begin{flalign}\label{eq:update}
\begin{split}
&V_{n}(t+1)=V_{n}(t)+a_{n}(t)*\Delta T\\
&\Delta V_{n-1,n}(t+1)=V_{n-1}(t+1)-V_{n}(t+1)\\
&S_{n-1,n}(t+1)=S_{n-1,n}(t)+\frac{\Delta V_{n-1,n}(t)+\Delta V_{n-1,n}(t+1)}{2}*\Delta T
\end{split}
\end{flalign}
where $\Delta T$ is the simulation time interval, set as $0.1s$ in this study, and $V_{n-1}$ is the velocity of lead vehicle (LV), which was externally inputted.

\subsection{Simulation Setup}To enable the RL agent to learn from trial and error, a simple car-following simulation environment was implemented. Initialized with the empirically given following vehicle speed, spacing and velocity differences,$V _ { n } ( t = 0 ) = V _ { n } ^ { data } ( t = 0 ) , S _ { n - 1 , n } ( t = 0 ) = S _ { n - 1 , n } ^ { d a t a } ( t = 0 ) , \text {and } \Delta V _ { n - 1 , n } ( t = 0 ) = \Delta V _ { n - 1 , n } ^ { d a t a } ( t = 0 )$, the RL agent is used to compute the acceleration $a_n(t)$. Given acceleration, future FV velocity, relative speed, and spacing are then generated iteratively based on (\ref{eq:update}). Once a car-following event reaches its ending, the state is re-initialized with empirical data of the next event.

\subsection{Reward function}The reward function, \textit{r}(\textit{s}, \textit{a}), serves as a training signal to encourage or discourage behaviors in the context of a desired task. For the task of autonomous car following, a reward function was established based on a linear combination of the features constructed in section \ref{feature}:
\begin{equation}\label{eq:totalfeature}
    r = w _ { 1 } F _ { T T C } + w _ { 2 } F _ { h e a d w a y } - w _ { 3 } F _ { j e r k }
\end{equation}
where $w_1, w_2, \text{and} w_3$  are coefficients of the features, all set as 1 in the current study.

\subsection{Network Architecture}The actor and critic was each represented by a neural network. The input of the actor network is the state at time step t, $s_t = (V_n(t), \Delta V_{n-1,n}(t), S_{n-1,n} (t))$. Its output is FV's acceleration $a_n(t)$. The input of the critic network is a state-action pair $(s_t,a_t)$. Its output is a scalar \textit{Q}-value $Q(s_t,a_t)$. 

\begin{figure}[h]
    \centering
    \includegraphics[width=0.48\textwidth]{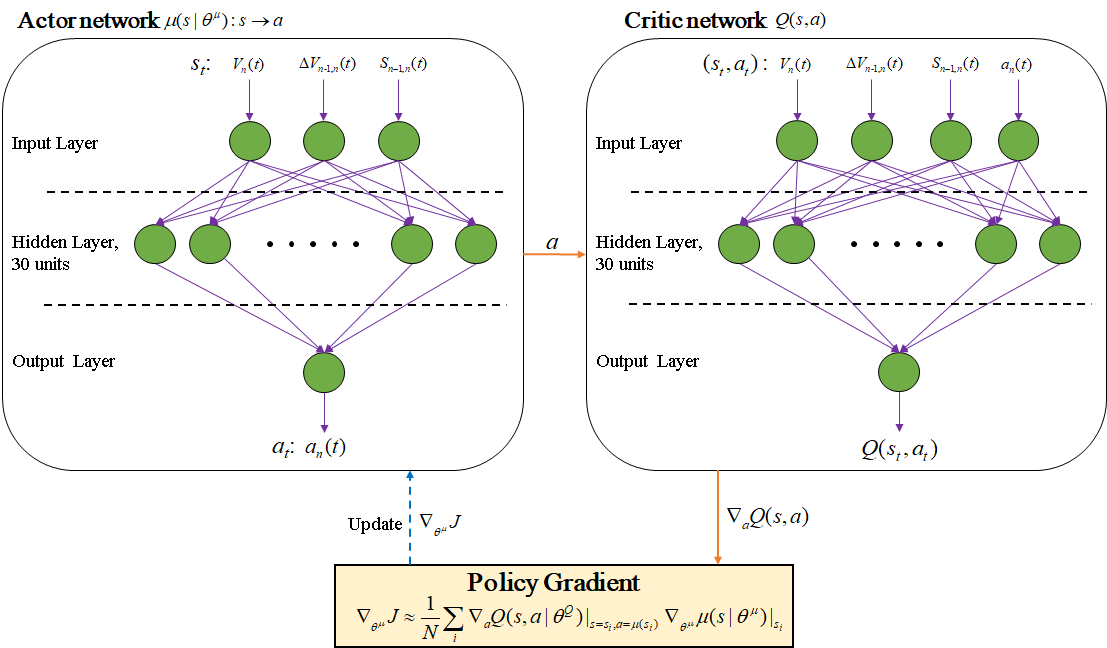}
    \caption{Architecture of the actor and critic networks.}
    \label{fig:network}
\end{figure}

Fig. \ref{fig:network} presents the architectures of the actor and critic networks \cite{zhu2018human}. Both of them consist of three layers: an input layer, an output layer, and a hidden layer with 30 neurons. Deeper neural networks with more than one hidden layers were also tested, but the results showed that they did not perform significantly better.

For the hidden layers, the Rectified Linear Unit (ReLU) activation function (\textit{f}(\textit{x}) = max(0, \textit{x})) was used. The ReLU can accelerate the convergence of network parameter optimization\unskip~\cite{369980:8164272}. For the output layer of the actor network, a $tanh$ activation function was used. The $tanh$ function maps real-valued numbers to the range [-1, 1] and thus can bound the outputted accelerations between -3 to 3 m/s\ensuremath{^{2}}.

\subsection{Network Update and Hyper Parameters}The parameters of the networks were updated based on Adam\unskip~\cite{369980:8164273} optimization algorithm. The critic network was updated by minimize the loss function $L = \frac{1}{N}\sum\nolimits_i {{{({y_i} - Q({s_i},{a_i}|{\theta ^Q}))}^2}} $; the actor network was updated according to the gradient ${\nabla _{{\theta ^\mu }}}J \approx \frac{1}{N}\sum\limits_i {{\nabla _a}Q(s,a|{\theta ^Q}){|_{s = {s_i},a = \mu ({s_i})}}{\nabla _{{\theta ^\mu }}}\mu (s|{\theta ^\mu })} {|_{{s_i}}}$ \cite{369980:8164262}.

The hyperparameters (parameters set prior to the training process) adopted are presented in Table I, these values were determined according to Lillicrap et al.\unskip~\cite{369980:8164262}  and also by performing a test on a randomly sampled training dataset.

\begin{table}[!htbp]\label{table:hyper}
\caption{{Hyperparameters and Corresponding Descriptions}}
\label{table-wrap-13b8985d42ecde88bc4004a354d4d958}
\def\arraystretch{1}
\ignorespaces 
\centering 
\begin{tabulary}{\linewidth}{p{\dimexpr.33\linewidth-2\tabcolsep}p{\dimexpr.33\linewidth-2\tabcolsep}p{\dimexpr.34\linewidth-2\tabcolsep}}
\hline \hline
Hyperparameter &
  Value &
  Description\\
\hline
Learning rate &
  0.001 &
  The learning rate used by Adam\\
Discount factor &
  0.99 &
  Discount factor gamma used in the Q-learning update\\
Minibatch size &
  32 &
  Number of training cases over which each stochastic gradient descent (SGD) update is computed\\
Replay memory size  &
  7000 &
  Number of training samples in the replay memory\\
Soft target update $\tau$ &
  0.001 &
  The update rate of target networks \\
\hline \hline
\end{tabulary}\par 
\end{table}

\subsection{Exploration Noise of Action}An exploration policy was constructed by adding noise sampled from a noise process to the original actor policy. as suggested by Lillicrap et al.\unskip~\cite{369980:8164262}, an Ornstein-Uhlenbeck process\unskip~\cite{369980:8164274}  with \textit{\ensuremath{\theta } } = 0.15 and \textit{\ensuremath{\sigma  }} = 0.2 was used. The Ornstein-Uhlenbeck process models the velocity of a Brownian particle with friction, generating temporally correlated values centered around zero. The temporally correlated noise enables the agent to explore well in a physical environment that has momentum.

\subsection{Training the DDPG Velocity Control Model}For the 1,341 extracted car-following events, 70\% (938) were used for training, and 30\% were used for testing. At the training stage, the RL agent sequentially simulates all the car-following events in the training data. Whenever a car-following event terminates and a new event is to be simulated, the state of the agent is initialized with the empirical data of the new one.

\begin{figure}[h]
    \centering
    \includegraphics[width=0.5\textwidth]{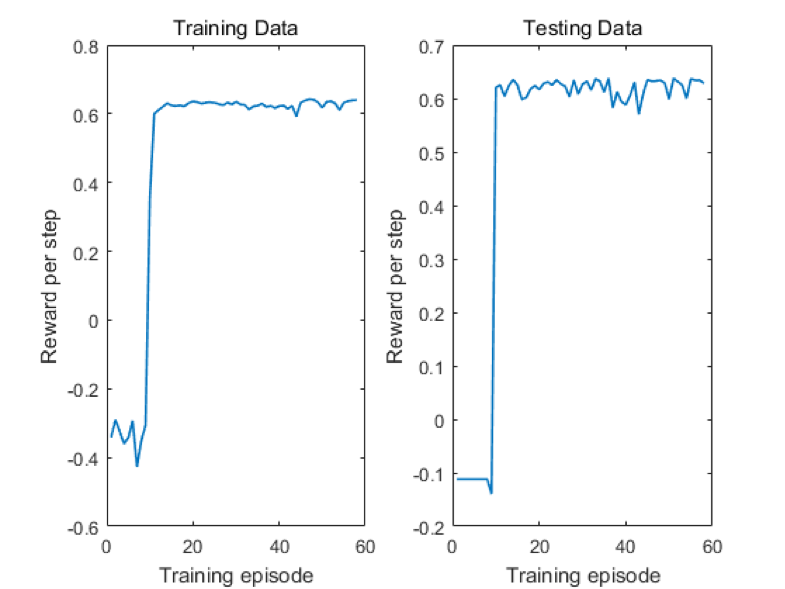}
    \caption{Reward curves during training.}
    \label{fig:train}
\end{figure}

The training was repeated for 60 episodes, and the RL agent generated the maximum average step reward on testing data was selected. Fig. \ref{fig:train} shows the change of average step reward with respect to training episode. As can be seen, the performance of the DDPG model starts to converge when the training episode reaches 20. When the model converges, the agent receives a reward value of about 0.64. This is achieved by selecting actions in a way that makes TTC and jerk feature values near 0 and get maximum headway features (0.65). It should be noted that the model has similar performances on training and testing data, demonstrating that it can generalize well on new data.

\section{RESULTS}
In this section, car-following behavior observed in the empirical NGSIM data and that simulated by the DDPG model were compared, to demonstrate the model's ability to follow a leading vehicle safely, efficiently, and comfortably. The DDPG model produces the following vehicle trajectories by taking the leading vehicle trajectories as input.

\subsection{Safe Driving}
\begin{figure}[h]
    \centering
    \includegraphics[width=0.5\textwidth]{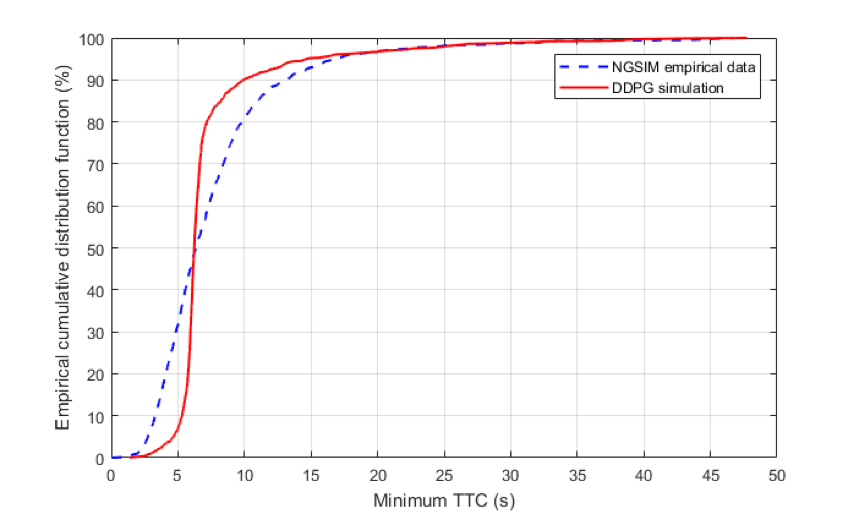}
    \caption{Cumulative distribution of minimum TTC during car following.}
    \label{fig:ttcres}
\end{figure}

Driving safety is evaluated based on minimum TTC during a car-following event. Fig. \ref{fig:ttcres} shows the cumulative distributions of minimum TTC for NGSIM empirical data and DDPG simulation. Nearly 35\% of NGSIM minimum TTCs were lower than 5s, while only about 8\% of DDPG minimum TTCs were lower than 5s. This means that car-following behavior generated by DDPG model is much safer than drivers' behavior observed in the NGSIM data.

\begin{figure}[h]
    \centering
    \includegraphics[width=0.5\textwidth]{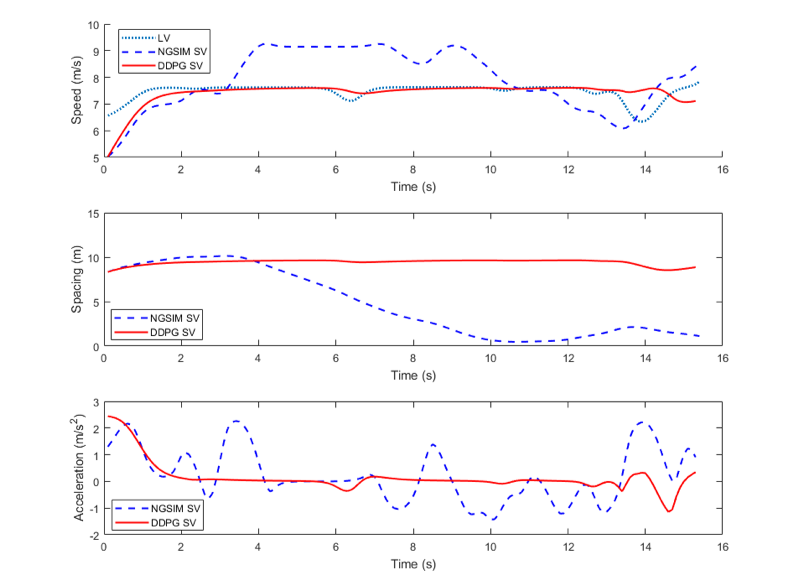}
    \caption{Comparison of driving safety between NGSIM data and the DDPG model.}
    \label{fig:safe}
\end{figure}

To give an illustration of the safe driving of the DDPG model, a car-following event was randomly chosen from the NGSIM dataset. Fig. \ref{fig:safe} shows the observed speed, spacing, and acceleration, and the corresponding ones generated by the DDPG model. The driver in the NGSIM data drove in a way that produced very small inter-vehicle spacing, while the DDPG model keeps a safe following gap around 10m.

\subsection{Efficient Driving}
\begin{figure}[h]
    \centering
    \includegraphics[width=0.5\textwidth]{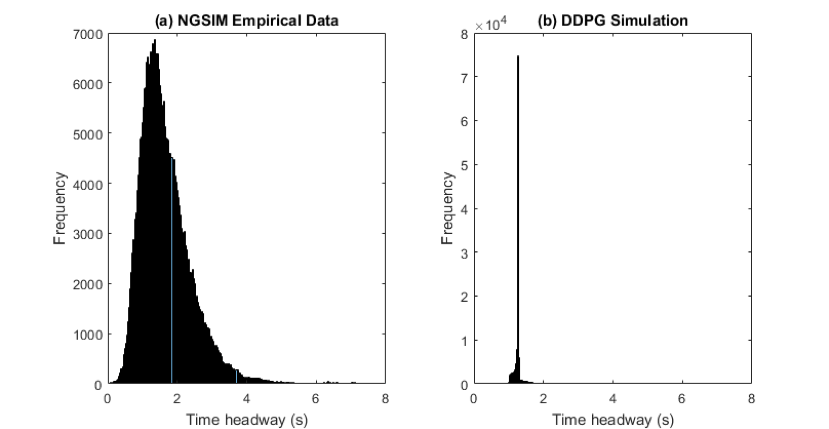}
    \caption{Histograms of time headway during car following for (a) NGSIM empirical data and (b) DDPG simulation.}
    \label{fig:hdwres}
\end{figure}

Time headway during car-following process was used to evaluate driving efficiency. Time headway was calculated at every time step of a car-following event, and the distribution of these time headways is shown in Fig. \ref{fig:hdwres}. 

As can been seen, the DDPG model produced car-following trajectories that always maintained a time headway in the range of 1s to 2s. While the NGSIM data had a much wider range of time headway distribution (0s to 6s). This included some dangerous headways that were less than 1s, and also some inefficient headways that were larger than 3s. Therefore, d it can be concluded that the DDPG model has the ability to follow the leading vehicle with an efficient and safe time headway.

\subsection{Comfortable Driving}
\begin{figure}[h]
    \centering
    \includegraphics[width=0.5\textwidth]{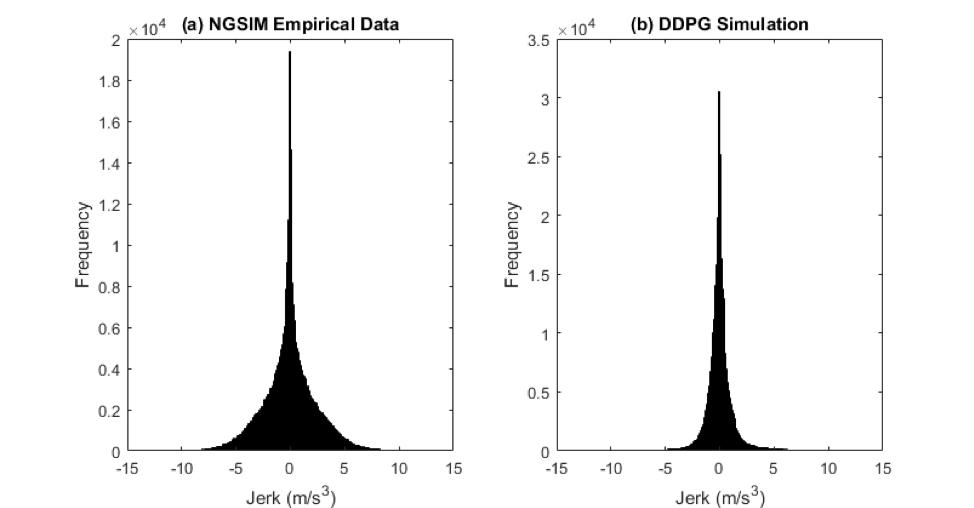}
    \caption{Histograms of jerk during car following for (a) NGSIM empirical data and (b) DDPG simulation.}
    \label{fig:jerkres}
\end{figure}

Driving comfort was evaluated based on jerk values during car following. Similar to time headway, it was calculated for every time step of a car-following event. Fig. \ref{fig:jerkres} presents the histograms of jerk values during car following. 

It is obvious that the DDPG model produced trajectories with lower values of jerk. Firstly, the DDPG trajectories had a narrow jerk distribution range (\mbox{\textminus}5 to 5 m/s\ensuremath{^{3}}) than NGSIM data (\mbox{\textminus}10 to 10 m/s\ensuremath{^{3}}). Second, jerk values were centered more closely to zero in DDPG simulation trajectories than in NGSIM data. As smaller absolute values of jerk correspond to more comfortable driving, it can be concluded that the DDPG model can control vehicle velocity in a more comfortable way than human drivers in the NGSIM data.

\begin{figure}[h]
    \centering
    \includegraphics[width=0.5\textwidth]{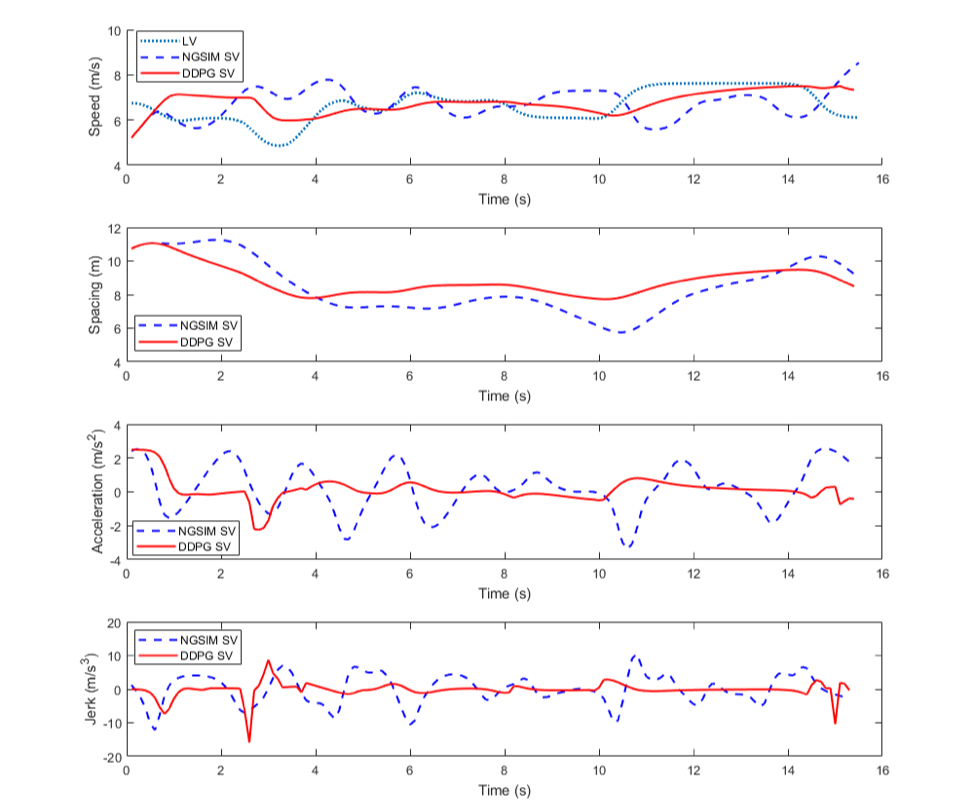}
    \caption{Comparison of driving comfort between NGSIM data and the DDPG model.}
    \label{fig:jerkdemo}
\end{figure}

To give an illustration of the comfortable driving of the DDPG model, a car-following event was randomly chosen in the NGSIM dataset. Fig. \ref{fig:jerkdemo} shows the observed speed, spacing, acceleration, and jerk, and the corresponding ones generated by the DDPG model. The driver in the NGSIM data drove in a way with frequent acceleration changes and large jerk values, while the DDPG model can remain a nearly constant acceleration and produced low jerk values. 

To summarize, the DDPG model demonstrated the capability of safe, efficient, and comfortable driving in that it 1) had small percentages of dangerous minimum TTC values that is less than 5 seconds; 2) could maintain efficient and safe headways within the range of 1s to 2s; and 3) followed the leading vehicle comfortably with smooth acceleration.
    
\section{DISCUSSION AND CONCLUSION}
A model used for velocity control during autonomous car-following was proposed based on deep RL. The model uses deep deterministic policy gradient (DDPG) algorithm to learn from trials and interaction, with a reward function signaling how the RL agent performs. The reward function was developed by referencing human driving data and combining driving features related to safety, efficiency, and comfort. By doing this, the model not only tries to imitate human drivers' behavior but also directly optimizes  driving  safety,  efficiency,and  comfort. Results show that compared to human drivers in the real world, the proposed DDPG car-following model demonstrated a better capability of safe, efficient, and comfortable driving. 

The proposed model can be further extended in the following aspects:

  \begin{enumerate}
  \item \relax More objectives can be added, such as energy-saving driving;
  \item \relax The weights of the objectives can be adjusted to reflect users' individual preferences;
  \item \relax In the current, a linear function was used to combine different objectives (features). More complicated reward function forms can be adopted to express more complex reward mechanisms, such as a non-linear function. 
  \end{enumerate}
  
  Although this study used networks with only one hidden layer, the key idea of deep reinforcement learning methods is fully exploited. Moreover, the networks can be easily extended to deeper ones once more input variables are provided.

This study can further be improved by designing better experience replay mechanisms. Experience replay lets RL agents remember and reuse experiences from the past. In the currently adopted DDPG algorithm, experience transitions were uniformly sampled, without considering their significance\unskip~\cite{369980:8164275}. In future work, prioritizing experience can be utilized to replay important transitions more frequently, and therefore learn in a more efficient way. 

To sum up, this study uses deep RL to learn how to control vehicle velocity during car following in a safe, efficient, and comfortable way.  Human driving data of the real world, from the NGSIM study, was used to train the model. Car-following behavior produced by the model were, then, compared with the observed one in the empirical NGSIM data, to evaluate the model's performance. Results show that the proposed model demonstrated the capability of safe, efficient, and comfortable driving, and may even perform better than human drivers. The results indicate that reinforcement learning methods could contribute to the development of autonomous driving systems. 

ACKNOWLEDGEMENTS

This study was sponsored by the Chinese National Science Foundation (51522810).



%

\bibliographystyle{IEEEtran}

\bibliography{article}

\end{document}